\documentclass[conference]{IEEEtran}
\usepackage{times}

\usepackage[numbers]{natbib}
\usepackage{multicol}
\usepackage[bookmarks=true]{hyperref}
\usepackage{amsmath} 
\usepackage{amssymb} 
\usepackage{algpseudocode}
\usepackage{xspace} 
\usepackage{graphicx}
\usepackage{stfloats}

\begin{document}

\title{CAMON: Cooperative Agents for Multi-Object Navigation with LLM-based Conversations}

\newcommand{\ours}{\textbf{\textsc{\textit{C{\small AMON}}}}\xspace}

\author{Pengying Wu, Yao Mu, Kangjie Zhou, Ji Ma, Junting Chen, Chang Liu}

\maketitle

\begin{abstract}
Visual navigation tasks are critical for household service robots. As these tasks become increasingly complex, effective communication and collaboration among multiple robots become imperative to ensure successful completion. In recent years, large language models (LLMs) have exhibited remarkable comprehension and planning abilities in the context of embodied agents. However, their application in household scenarios, specifically in the use of multiple agents collaborating to complete complex navigation tasks through communication, remains unexplored. Therefore, this paper proposes a framework for decentralized multi-agent navigation, leveraging LLM-enabled communication and collaboration. By designing the communication-triggered dynamic leadership organization structure, we achieve faster team consensus with fewer communication instances, leading to better navigation effectiveness and collaborative exploration efficiency. 
With the proposed novel communication scheme, our framework promises to be conflict-free and robust in multi-object navigation tasks, even when there is a surge in team size.
\end{abstract}

\IEEEpeerreviewmaketitle

\section{Introduction}

\begin{figure*}[ht]
  \centering

   \includegraphics[width=0.9\linewidth]{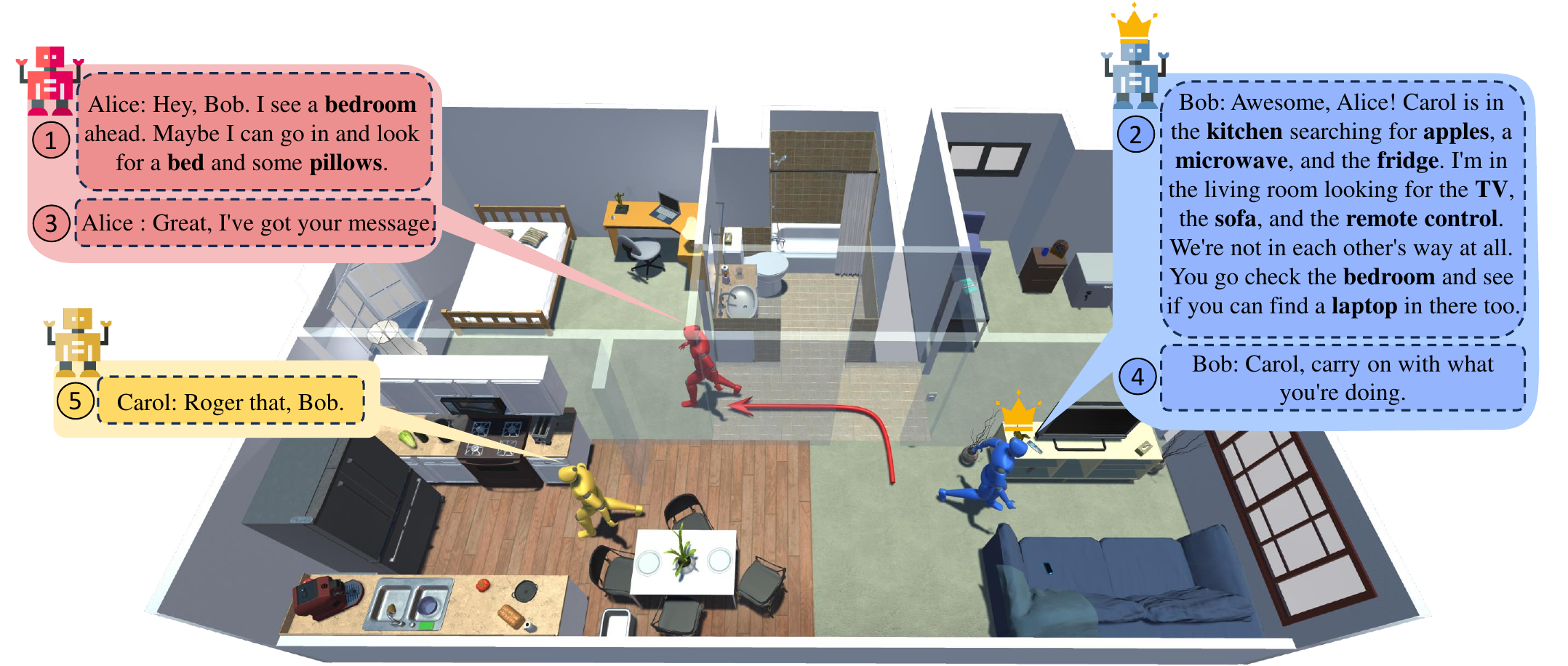}

   \caption{We contribute \ours: a framework for Cooperative Multi-Object Navigation in indoor Environments. This figure shows three agents collaborating to find some objects, and the dialog box represents the agents' conversation contents. In \ours, the agents make decisions that do not conflict with other robots and maximize team collaboration benefits by asking their current leaders.}
   \label{fig:teaser}
   \vspace{-5pt}
\end{figure*}

In recent years, household visual navigation utilizing the large language model (LLM) has advanced rapidly. 
Previous methods \citep{zhou2023esc, 10373065, shah2023lfg, cai2023bridging, chen2024mapgpt} have leveraged LLMs as scene-understanding tools and planners, yielding promising application results. 
However, these approaches are constrained to single-agent navigation and do not offer viable solutions for effective communication and collaborative planning among multiple agents \citep{yu2023co}. 
When tasked with searching for and locating various objects within a household environment, such complex tasks pose significant challenges for a single robot, leading to low efficiency and high failure rates. 
In multi-object navigation scenarios \citep{wani2020multion,marza2023multi,gireesh2023sequence}, it is essential for multiple robots to collaborate to accomplish these tasks effectively.

Successfully completing multi-agent tasks necessitates a team possessing three key abilities:
(1) extracting useful information from observations, i.e., \textit{determining the content of communication}, 
(2) a conflict-free communication mechanism, i.e., \textit{identifying with whom to communicate}, 
(3) a global planning capability, i.e., \textit{planning after communication}.
To achieve these abilities, we have designed a novel framework specifically designed for multi-agent navigation, and the effect is shown in Figure \ref{fig:teaser}. This method achieves cooperative multi-target tasks through structured scene descriptions and ordered communication mechanisms.

To logically organize and summarize observations, we focus on the layout patterns of indoor scenes, where the placement of objects in household environments is often related to the properties of the room \citep{sun2024leveraging}. 
For example, a room with a bed is typically a bedroom, where pillows, televisions, and similar items are commonly found, while toilets and microwaves are not. 
Therefore, we advocate this motivation of object-room relationships in navigation representations, dividing the observed scene into individual rooms and generating descriptions of each room for subsequent communication for task division. 
For instance, upon entering a room and identifying it as a living area based on its layout, a robot should promptly locate and exclusively find all potential targets within that space. 
To ensure efficient team collaboration, other robots should avoid entering rooms identified as living areas upon detection.

In multi-agent embodied tasks, the organizational structure of the team is crucial \citep{chen2023multi, guo2024embodied}. 
Prior research \citep{chen2023scalable} indicates that leadership-based communication patterns expedite consensus achievement, whereas dynamic leadership allocation further enhances team coordination and effectiveness \citep{guo2024embodied}. 
To establish an efficient and stable communication system, we designed a comm-triggered\footnote{"comm" is short for "communication" in this paper.} dynamic leadership model as an advanced form of decentralized communication. 
Our contributions can be summarized as follows \footnote{This work is in progress.}:

\begin{itemize}
\item We have designed a comprehensive framework for multi-agent navigation tasks utilizing LLMs, encompassing modules for perception, communication, and cooperative planning.

\item Our proposed multi-agent communication mechanism facilitates adaptive task division and planning for complex navigation tasks.

\item We have developed a dynamic leadership mechanism, activated by agents' communication requests, that facilitates the distribution of information exchange workload in distributed systems.

\end{itemize}

\section{Related Work}

\subsection{Visual Object Navigation}

Target object navigation necessitates that robots swiftly locate and approach the target object in an unfamiliar environment. 
Recent research in this field has primarily branched into two mainstream approaches: end-to-end network model-based frameworks \citep{anderson2018vision, majumdar2022zson, 10161345, 10161289, chen2023zero, gadre2023cows, cai2023bridging} and modular map-based frameworks \cite{ramakrishnan2022poni, chaplot2020object, chen2023not, zhou2023esc, yu2023l3mvn, shah2023lfg, wu2024voronav, sun2024leveraging, ma2024doze}. 
End-to-end model-based methods exhibit good transferability but are relatively ineffective in navigation efficiency and task success rates \cite{chaplot2020object}. 
Conversely, the modular approach, guided by hierarchical maps, requires meticulously designed modules, enabling highly effective navigation. 
With the increasing complexity of object-finding tasks, multi-object navigation (MultiON) tasks \citep{wani2020multion} and methods \citep{10161030, marza2023multi, 10100678} have emerged as advanced versions of single-target object navigation. 
However, existing methods for MultiON predominantly address pre-sequenced MultiON, where the robot receives a predefined sequence for exploring target object classes. 
To demonstrate the flexibility of the task and the adaptive planning capability of the proposed framework, we adopt the Sequence Agnostic MultiON (SAM) \citep{gireesh2023sequence} task for evaluation. 
In this approach, the robot neither receives nor is required to follow a global order for locating and navigating to instances of target object classes. 
Instead, the robot explores probable locations of the target objects and dynamically adapts its exploration based on observations.



\subsection{LLM-Based Cooperative Embodied Agents}

Recent work \citep{mandi2023roco, zhang2023building, zhang2023controlling} has demonstrated the feasibility of inputting observation in linguistic form into large language models for communication and decision-making in \textbf{M}ulti-\textbf{A}gent \textbf{S}ystems (MAS). 
Most of the work is structured in a hierarchical manner to ensure the proper functioning of MAS. 
The mainstream LLM-based multi-agent planning frameworks are divided into two major branches: centralized \citep{zhao2024hierarchical, agashe2023evaluating, yu2023co, chen2023scalable} and decentralized \citep{chen2023scalable, liu2023llm, zhang2023building, ying2024goma, wang2024safe}. 

In the centralized organization, the LLMs comprehend the observations, history, and task progress of multiple agents, and collaboratively allocate tasks to each robot group \citep{zhao2024hierarchical} or individual \citep{agashe2023evaluating, yu2023co}. 
Specifically, \citet{yu2023co} implement a centralized multi-agent navigation framework, extracting frontier information and semantic information from the map, and utilizing LLMs to allocate exploratory areas for each robot. 
Such frameworks achieve good coordination and planning performance in small-scale groups. 
However, as the team size scales up, the communication and information processing burden of centralized leadership increases, posing challenges to reasonable and seasonable planning \cite{chen2023scalable}. 

In decentralized systems, each robot acts as an independent entity with self-autonomy, exchanging historical observations through human-like verbal communication and making adaptive decisions \citep{ying2024goma}. 
In particular, \citet{zhang2023building} provide a systematic template for decentralized communication and collaboration. 
This method categorizes each agent's execution in the MAS into five modules: observation, belief, communication, reasoning, and planning, with the LLMs facilitating inter-agent communication and reasoning.

\subsection{ Multi-Agent Organizational Structures}

Recent studies have delved into the impact of organizational structures among multiple agents on task division, planning conflicts, and communication costs in embodied tasks. 
Through experiments, \citet{chen2023scalable} demonstrate that a hierarchical organizational structure with leadership significantly outperforms the original decentralized and centralized structures in terms of effectiveness. 
\citet{chen2023multi} also analyze the communication patterns among multiple agents and have validated that a leader-organized framework achieves faster task convergence through a simple task of multiple agents moving to a common point. 
\citet{guo2024embodied} further investigate the organizational forms of leaders and have concluded that dynamic leaders are the most effective in multi-agent collaboration. 
However, the scheme proposed by \citep{guo2024embodied} assigns leaders based on the importance of the tasks performed by each robot, which does not work well when tasks are equally important. 
Here, we propose a comm-triggered dynamic leader strategy that can perform well in MultiON tasks.

\section{Problem Setup}
We consider having multiple collaborative robots participate in MultiON tasks, where they jointly explore and approach target objects in indoor environments. 
At the beginning of each episode, two or more robots are randomly placed in an unfamiliar environment and tasked with searching for a common set of target objects $G = \{g_1, g_2, \ldots , g_m \}$. 
The robots have neither been trained to find these objects nor have prior information about similar scenes. They must coordinate their exploration and locate the targets as quickly as possible using general knowledge and common sense. 
Through communication, the robots can collaborate to understand the scene, divide tasks, and navigate to the target objects' locations efficiently. 

When a robot correctly recognizes and approaches a target object $g_t \in G$, the sub-task is considered successful. 
Conversely, if a robot navigates to the wrong object, the sub-task is considered a failure. 
Regardless of success or failure, the robot will continue searching for the remaining objects. 
Additionally, the episode ends if all robots reach the maximum step limit to avoid endless movement.

\begin{figure*}[t]
  \centering

   \includegraphics[width=0.9\linewidth]{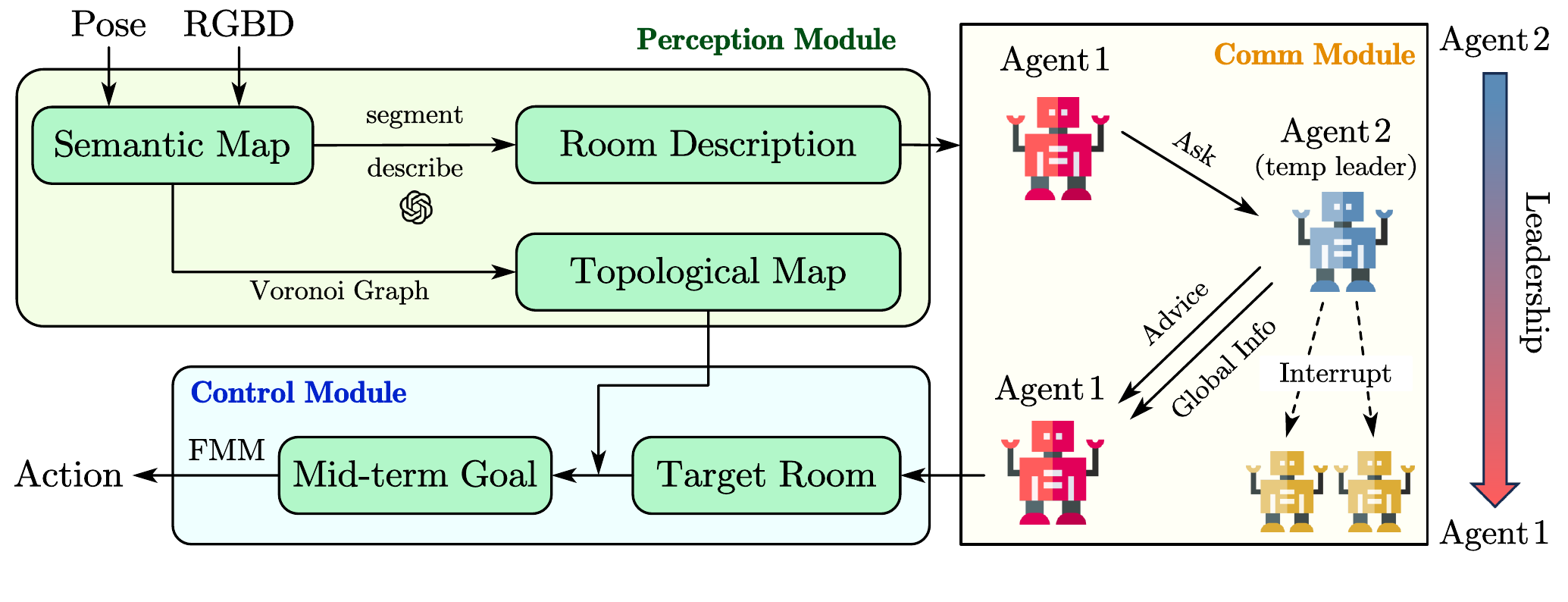}

   \caption{\textbf{Components of \ours}. Our framework comprises three modules: perception, communication, and control. The perception module generates a real-time semantic map using robot RGB-D and pose inputs, from which the agent extracts topology maps, and segments and describes rooms. Agent\_1 makes global decisions by querying the current leader, Agent\_2, to obtain the target room. Leadership and global information are then conveyed from Agent\_2 to Agent\_1. Finally, the control module generates a sequence of actions for the Agent\_1 to navigate from its current position to the target room.}
   \label{fig:model}
   \vspace{-5pt}
\end{figure*}

\section{\ours Approach}
The key concept of \ours is to present a comprehensive framework for completing multi-agent navigation tasks through LLM communication and planning. 
As shown in Figure \ref{fig:model}, the central technical approach is as follows: 
(1) In the perception module that understands and describes observed scenes (\ref{per}), each robot maintains a local map (\ref{per:map}), divides the map into room levels, extracts the most relevant historical image frames for each room, uses large multimodal models (GPT-4o) to interpret these image frames, and generates language descriptions of the observed scenes (\ref{per:roomdes}). 
(2) In the comm module that undertakes communication and decision-making (\ref{comm}), we employ a sequential and dynamic leader-member communication structure, where the leader adjusts the decision-making proposals (\ref{comm:ask}) of the agent and ensures team coordination (\ref{comm:coordination}). 
By making decisions during communication, the member will be assigned the responsibility of finding specific objects and will select the next referenced target room from the leader. 
(3) At the path planning level (\ref{motion}), we plan a sequence of waypoints on the map based on the current location and target room, generating discrete actions.

\subsection{Perception Module}
\label{per}
\subsubsection{Map Construction}
\label{per:map}
To record historical semantic information, each robot constructs and updates a local semantic map in real-time using its poses and RGB-D images. 
The map records the occupancy of obstacle areas, accessible areas, and semantic information. 
Inspired by the methodology in \citep{zuo2020improved,kwon2023renderable, wu2024voronav}, we extracted the waypoints and topological map from the accessible area map channels to optimize the point-to-point movement.

\subsubsection{Room Description}
\label{per:roomdes}

Following the principle of 3D room segmentation \citep{werby2024hierarchical, hughes2022hydra}, the observed rooms are segmented to obtain masks for each room on the map. 
To obtain a description of each room, we aim to use an LMM to read the holistic image of the entire room and generate the corresponding description. 
To capture such a comprehensive image, we take advantage of the robot's manner of moving between nodes on the topology map. 
The robot will rotate itself to collect $12$ frames of images each time it passes through waypoints. For each segmented room, we select the image from the recorded frames that best captures the view of the room.  
Next, we use GPT-4o \citep{GPT-4o} to generate descriptions of the scenes in these images.

\subsection{Planning Module}
\label{comm}

During navigation, each embodied agent carefully observes the current room and updates the semantic map of the room each time it enters a new one. 
When remaining and previously unsearched objects are detected, the agent sequentially navigates to the locations of the target objects. 
When the entire room has been explored and no remaining target objects are present in the room, the robot considers which room to explore next and what objects to take responsibility for by communicating with the current leader through a communication module to make a reasonable decision.

\subsubsection{Communication and Leadership Appointment}
\label{comm:information}
In team collaboration, the presence of leaders greatly affects communication efficiency and task completion. 
An orderly organization needs to address the question of who the leader is and what role the leader plays. 
In this framework, we adopt a comm-triggered dynamic leadership mechanism to answer the question of who is the leader, while addressing the issues of imbalanced information flow and poor robustness in fixed leadership. 
As the episode starts, one of the robots, $agent\_a$, collected preliminary room descriptions from all robots as global information, becoming a temporary leader. 
When another robot $agent\_b$ requests assistance from its leader in the future, it sends a request to $agent\_a$, who provides suggestions to $agent\_b$ and conveys the global information to $agent\_b$. 
At this time, $agent\_b$ updates the part of itself in the global information, and $agent\_b$ inherits the leader. 
In the subsequent process, robots continuously send communication requests, and global information and leadership are conveyed sequentially within the robots. 
We answer the question of the role played by leaders by endowing dynamic leaders with temporary access to the authority to command any robot and the latest global information. 
This collaborative approach can share the communication load between robots, and even if a robot crashes (or even the temp leader), the remaining robots can maintain system stability by asking the previous leader.

\subsubsection{Agent Asks for Help}
\label{comm:ask}

Due to the complexity of multi-agent task planning, LLM needs clear historical observations and team conditions to enhance understanding and planning performance. 
Before any robot team member sends a communication request to the current leader, a robot makes an initial decision using the LLM based on the recorded historical information. 
This decision focuses on the benefit of the robot itself, which is subsequently conveyed to the leader to help determine whether the decision will cause conflicts with other robots. 
When the robot receives the response from the leader containing the target room and locked objects to prevent others from finding them, it then moves to the target room. 
We use templates $Pr$ to concatenate agent id $i$, task progress $P_i$, recorded states $S_i$, shared goals $G$, conversation history $H_i$, and optional actions $\mathcal{A}_i$. 
Then the LLM generates the initial proposal $Ps_i = \{\mathcal{L}_i, a_i, T_i\} $, containing locked object $\mathcal{L}_i$, action $a_i$ (i.e., target room), and thoughts $T_i$. 
The process can be formulated as:

\begin{equation}
  Ps_i = \text{LLM}(Pr(i, P_i, S_i, G, H_i))
  \label{eq:member}
\end{equation}

\subsubsection{Leader's Coordination of Teamwork}
\label{comm:coordination}

When receiving requests from team members, the leader undertakes to evaluate the initial proposals $Ps_i$ from agent $i$ and avoid team conflicts based on the current recorded global states $S_g$ of all agents and task progress. 
Similarly, We employ the LLM to handle this process, as demonstrated in Formula \ref{eq:leader}.

\begin{equation}
\label{eq:leader}
  Re^* = \text{LLM}(Pr^*(Ps_i, P_g, S_g, G))
\end{equation}

where $Pr^*$ represents the leader's prompt template. 
The initial proposal from the $i$-th agent is denoted by $Ps_i$. 
The global task progress and global states, currently managed by the leader, are indicated by $P_g$ and $S_g$, respectively. 
The coordination result generated by LLM is $Re^* = \{\mathcal{R}_i, \ldots, \mathcal{R}_j\} $, where $R_i$ is the response to $i$-th agent who initiated the request, including the action $a_i^*$, assigned $D^*$ for supporting or opposing the original proposal from the team member, as well as thoughts $T_i^*$. 
The remaining responses $R_j \in Re^* \setminus R_i$ contain decisions on whether to interrupt $j$-th agent, as well as actions assigned to it (if interrupted).

\subsection{Motion Planning}
\label{motion}

Given the current position of the robot and the target room, the robot selects the Voronoi point closest to its current position within the target room. 
Subsequently, it uses the Dijkstra \citep{dijkstra1959note} method to plan the next waypoint along the topological edges as a mid-term goal. 
Then, the Fast Marching Method \citep{sethian1996fast} is employed to plan the shortest point sequence and the next discrete action in real-time, from the current position to the mid-term target point. 
Whenever the robot passes through a topological waypoint, it rotates once to collect surrounding images.



\section{Conclusion}
\label{sec:conclusion}

In this work, we propose a fully decentralized, LLM-based multi-agent collaborative navigation framework. 
The agents can effectively communicate, efficiently divide tasks and collaborate through a dynamic leadership organization. 
Our method can achieve task division and team planning consensus with minimal communication overhead, resulting in optimal navigation performance and robustness in the multi-agent system. 
We believe that our approach holds promising applications for future team collaboration of household mobile agents.

\textbf{Limitations \& Future Work.}
Several limitations remain in the current approach. 
First, while our mapping perception module can effectively map all observed objects, it may struggle with dynamic objects, such as humans or pets, in household scenarios. 
This could adversely affect the robot's mapping and room segmentation capabilities. 
Additionally, our current framework is restricted to single-floor navigation and does not account for cross-floor cooperative navigation. 
However, we believe these limitations are not central to the issues addressed in this article and can be mitigated by incorporating additional strategy modules. 
Our work demonstrates promising application prospects and offers feasible ideas for future multi-robot systems. 
Future research will focus on the collaboration of multi-robot system movement and manipulation, aiming to complete the framework for communication, navigation, and manipulation integration. 
These important topics will be left for future exploration.

\clearpage

\bibliographystyle{plainnat}
\bibliography{references}

\end{document}